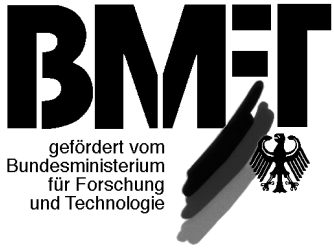
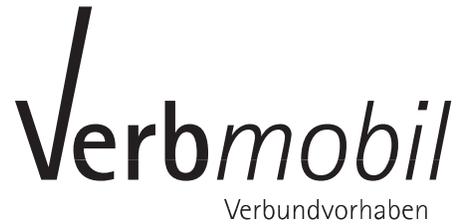

# A Robust and Efficient Three-Layered Dialogue Component for a Speech-to-Speech Translation System

Jan Alexandersson
Elisabeth Maier
Norbert Reithinger

DFKI GmbH

Report 50
Dezember 1994

Jan Alexandersson
Elisabeth Maier
Norbert Reithinger

DFKI GmbH
Stuhlsatzenhausweg 3
66123 Saarbrücken

Tel.: (0681) 302 - 5347, 5346
e-mail: {alexandersson}@dfki.uni-sb.de

# A Robust and Efficient Three-Layered Dialogue Component for a Speech-to-Speech Translation System

Jan Alexandersson   Elisabeth Maier
Norbert Reithinger *
DFKI GmbH, Stuhlsatzenhausweg 3
D-66123 Saarbrücken, Germany
{alexandersson,maier,reithinger}@dfki.uni-sb.de

**Abstract**

We present the dialogue component of the speech-to-speech translation system VERBMOBIL. In contrast to conventional dialogue systems it mediates the dialogue while processing maximally 50% of the dialogue in depth. Special requirements (robustness and efficiency) lead to a 3-layered hybrid architecture for the dialogue module, using statistics, an automaton and a planner. A dialogue memory is constructed incrementally.

## 1 Introduction

VERBMOBIL combines the two key technologies speech processing and machine translation. The long-term goal of this project is the development of a prototype for the translation of spoken dialogues between two persons who want to find a date for a business meeting (for more detail on the objectives of VERBMOBIL see [Wahlster, 1993]). A special characteristic of VERBMOBIL

---

is that both participants are assumed to have at least a passive knowledge of English which is used as intermediate language. Translations are produced *on demand* so that only parts of the dialogue are processed. If VERBMOBIL is inactive, shallow processing by a keyword spotter takes place which allows the system to follow the dialogue at least partially.

In this paper focus is on the description of the dialogue component, which processes the interaction of the two dialogue partners and builds a representation of the discourse. Dialogue processing in VERBMOBIL differs from systems like SUNDIAL [Andry, 1992] in two important points: (1) VERBMOBIL *mediates* the dialogue between two human dialogue participants; the system is not a participant of its own, i.e. it does not control the dialogue as it happens in the flight scheduling scenario of SUNDIAL; (2) VERBMOBIL processes maximally 50% of the dialogue contributions in depth, i.e. when the 'owner' of VERBMOBIL speaks German only. The rest of the dialogue can only be followed by a keyword spotter.

In the remainder of this paper first the requirements of the VERBMOBIL setting with respect to functionality and design of the dialogue component section are introduced. Then a hybrid architecture for the dialogue component and its embedding into the VERBMOBIL prototype are discussed. Finally, results from our implemented system are presented. We conclude with an outline of future extensions.

## 2  Tasks of the Dialogue Component

The dialogue component within VERBMOBIL has four major tasks:

(1) to support speech recognition and linguistic analysis when processing the speech signal. *Top-down predictions* can be made to restrict the search space of other analysis components to get better results in shorter time [Young et al., 1989, Andry, 1992]). For instance, predictions about a speech act can be used to narrow down the set of words which are likely to occur in the following utterance – a fact exploited by the speech recognition component which uses adaptive language models [Jellinek, 1990]. Top-down predictions are also used to limit the set of applicable grammar rules to a specific subgrammar. They are of particular importance since the system has to work under real-time constraints.

(2) to provide contextual information for other VERBMOBIL components. In order to get good translations, context plays an important role. One example is the translation of the German *"Geht es bei Ihnen?"* which can be translated as *"Does it suit you?"* or *"How about your place?"*, depending on whether the dialogue partners discussed a time or a place before. A discourse history is constructed which can be accessed by other VERBMOBIL components for making contextual inferences [Ripplinger and Caroli, 1994, LuperFoy and Rich, 1992].

(3) to follow the dialogue when VERBMOBIL is off-line. When both dialogue participants speak English (and no automatic translation is necessary) VERBMOBIL is "passive", i.e. no syntactic or semantic analyses are performed. In such cases, the dialogue component tries to follow the dialogue by using a keyword spotter. This device scans the input for a small set of predetermined words which are characteristic for certain stages of the dialogue. The dialogue component computes the most probable speech act type of the next utterance which is used to selects its typical key words.

(4) to control clarification dialogues between VERBMOBIL and its users. If processing breaks down VERBMOBIL has to initiate a clarification dialogue in order to recover.

# 3 The Architecture

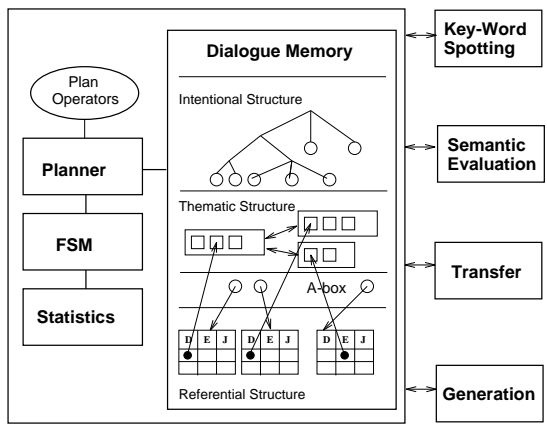

Figure 1: Architecture of the dialogue module

The abovementioned requirements cannot be met when using a single method of processing: if we use structural knowledge sources like plans or dialogue-grammars, top-down predictions are difficult make, because usually one can infer many possible follow-up speech acts from such knowledge sources that are not scored [Nagata and Morimoto, 1993]. Also, a planning-only approach is inappropriate when the dialogue is processed only partially. Therefore we chose a hybrid 3-layered approach (see fig. 1) where the layers differ with respect to the type of knowledge they use and the task they are responsible for. The processing components are

**A Statistic Module** The task of the statistic module is the prediction of the following speech act (see section 4), using knowledge about speech act frequencies in our training corpus.

**A Finite State Machine (FSM)** The finite state machine describes the sequence of speech acts that are admissible in a standard appointment scheduling dialogue and checks the ongoing dialogue whether it follows these expectations (see fig. 2).

**A Planner** The hierarchical planner constructs a description of the dialogue's underlying dialogue and thematic structures, making extensive use of contextual knowledge. This module is sensitive to inconsistencies and therefore robustness and backup-strategies are the most important features of this component.

While the statistical component completely relies on numerical information and is able to provide scored predictions in a fast and efficient way, the planner handles time-intensive tasks exploiting various knowledge sources, in particular linguistic information. The FSM can be located in between these two components: it works like an efficient parser for the detection of inconsistent dialogue states. The three modules interact in cases of repair, e.g. when the planner needs statistical information to resume an incongruent dialogue.
On the input side the dialogue component is interfaced with the output from the *semantic construction/evaluation* module, which is a DRS-like feature-value structure [Bos et al., 1994] containing syntactic, semantic, and occasionally pragmatic information. The input also includes information from the generation component about the utterance produced in the target language and a word lattice from the *keyword spotter*.
The output of the dialogue module is delivered to any module that needs information about the dialogue pursued so far, as for example the transfer module and the semantic construction/evaluation module. Additionally, the keyword spotter is provided with words expected in the next utterance.

## 4 Layered Dialogue Processing

### 4.1 Knowledge-Based Layers

#### 4.1.1 The Underlying Knowledge Source – The Dialogue Model

Like previous approaches for modeling task-oriented dialogues we base our ideas on the assumption that a dialogue can be described by means of a limited but open set of speech acts (see e.g. [Bilange, 1991], [Mast, 1993]). As point of departure we take *speech acts* as proposed by [Austin, 1962] and [Searle, 1969] and also a number of so-called *illocutionary acts* as employed in a model of information-seeking dialogues [Sitter and Stein, 1992]. We examined the VERBMOBIL corpus of appointment scheduling dialogues for their occurrence and for the necessity to introduce new speech acts[1].

At present, our model contains 17 speech acts (see [Maier, 1994] for more details on the characterization of the various speech acts; the dialogue model describing admissible sequences of speech acts is given in fig. 2). Among the domain-dependent speech acts there are low-level (primitive) speech acts like BEGRUESSUNG for initiating and VERABSCHIEDUNG for concluding a dialogue. Among the domain-independent speech acts we use acts as e.g. AKZEPTANZ and ABLEHNUNG. Additionally, we introduced two speech acts necessary for modeling our appointment scheduling dialogues: INIT_TERMINABSPRACHE and BESTAETIGUNG. While the first is used to describe utterances which state dates or places to be negotiated, the latter corresponds to contributions that contain a mutual agreement concerning a given topic.

The dialogue consists of three phases [Maier, 1994]. First, an introductory phase, where the discourse participants greet each other, introduce themselves and provide information e.g. about their professional status. After this, the topic of the conversation is introduced, usually the fact that one or more appointments have to be scheduled. Then negotiation begins where the discourse participants repeatedly offer possible time frames, make counter of-fers, refine the time frames, reject offers and request other possibilities. Once an item is accepted and mutual agreement exists either the dialogue can be terminated. or another appointment is negotiated.

---

[1] The acts we introduce below are mostly of illocutionary nature. Nevertheless we will refer to them as *speech acts* throughout this paper.

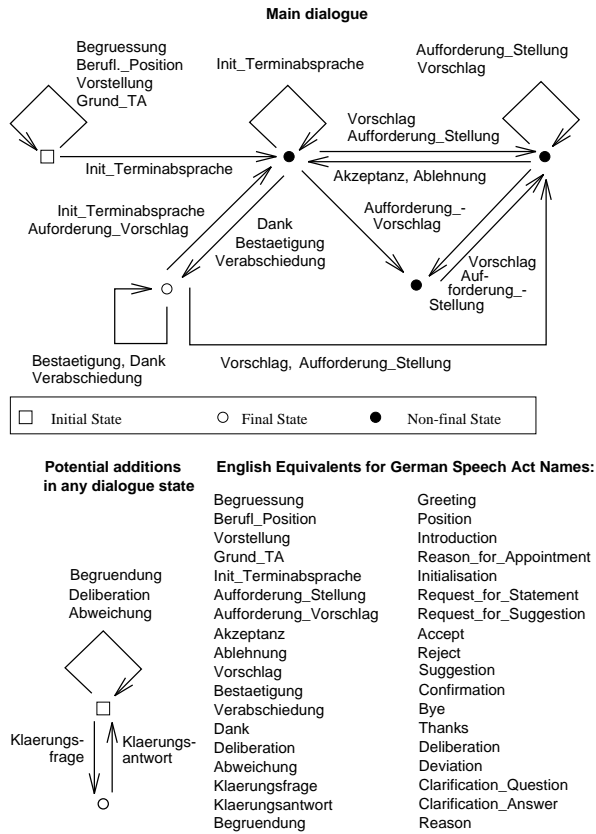

Figure 2: A dialogue model for the description of appointment scheduling dialogues

A dialogue model based on speech acts seems to be an appropriate approach also from the point of view of machine translation and of transfer in particular: While in written discourse sentences can be considered the basic units of transfer, this assumption is not valid for spoken dialogues. In many cases only sentence fragments are uttered, which often are grammatically incomplete or even incorrect. Therefore different descriptive units have to be chosen. In the case of VERBMOBIL these units are speech acts.

The speech acts which in our approach are embedded in a sequential model of interaction can be additionally classified using the taxonomy of dialogue control functions as proposed in e.g. [Bunt, 1989]. Speech acts like BEGRUES-SUNG and VERABSCHIEDUNG, for example, can be classified as dialogue functions controlling interaction management. More fine-grained taxonomical distinctions like CONFIRM and CONFIRM/WEAK as proposed in [Bunt, 1994] are captured in our approach by pragmatic features like *suitability* and *possibility* specified in the DRS-description of an utterance, which serves as input for the dialogue component.

### 4.1.2 The Finite State Machine

The finite state machine provides an efficient and robust implementation of the dialogue model. It parses the speech acts encountered so far, tests their consistency with the dialogue model and saves the current state. When an inconsistency occurs fall back strategies (using for instance the statistical layer) are used to select the most probable state. The state machine is extended to allow for phenomena that might appear anywhere in a dialogue, e.g. human-human clarification dialogues and deliberation. It can also handle recursively embedded clarification dialogues.
An important task of this layer is to signal to the planner when an inconsistency has occurred, i.e. when a speech act is not within the standard model so that it can activate repair techniques.

### 4.1.3 The Dialogue Planner

To incorporate constraints in dialogue processing and to allow decisions to trigger follow-up actions a plan-based approach has been chosen. This approach is adopted from text generation where plan-operators are responsible for choosing linguistic means in order to create coherent stretches of text (see, for instance, [Moore and Paris, 1989] and [Hovy, 1988]). The application of plan operators depends on the validity of constraints. Planning proceeds in a top-down fashion, i.e. high-level goals are decomposed into *subgoals*, each of which has to be achieved individually in order to be fulfilled. Our top-level goal `SCHEDULE-MEETING` (see below) is decomposed into three subgoals each of which is responsible for the treatment of one dialogue segment: the introductory phase (`GREET-INTRODUCE-TOPIC`), the negotiation phase (`NEGOTIATE`) and the closing phase (`FINISH`). These goals have to be fulfilled in the specified order. The keyword `iterate` also specifies that
negotiation phases can occur repeatedly.

```
begin-plan-operator GENERIC-OPERATOR
   goal [SCHEDULE-MEETING]
   constraints nil
   actions nil
   subgoals (sequence [GREET-INTRODUCE-TOPIC]
                      iterate [NEGOTIATE]
                      [FINISH])
end-plan-operator

begin-plan-operator OFFER-OPERATOR
   goal [OFFER]
   constraints nil
   actions (retrieve-theme)
  subgoals primitive
end-plan-operator
```

In our hierarchy of plan operators the leaves, i.e. the most specific operators, correspond to the individual speech acts of the model as given in fig. 2. Their application is mainly controlled by pragmatic and contextual constraints. Among these constraints are, for example, features related to the discourse participants (acquaintance, level of expertise) and features related to the dialogue history (e.g. the occurrence of a certain speech act in the preceding context).

Additionally, our plan operators contain an *actions* slot, where operations which are triggered after a successful fulfillment of the subgoals are specified. Actions, therefore, are employed to interact with other system components. In the sub-plan OFFER-OPERATOR, for example, which is responsible for planning a speech act of the type VORSCHLAG, the action (retrieve-theme) filters the information relevant for the progress of the negotiation (e.g. information related to dates, like months, weeks, days) and updates the thematic structure of the dialogue history. During the planning process tree-like structures are built automatically which mirror the structure of the dialogue.

The dialogue memory consists of three layers of dialog structure: (1) an intentional structure representing dialogue phases and speech acts as occurring in the dialogue, (2) a thematic structure representing the dates being negotiated, and (3) a referential structure keeping track of lexical realizations. The planner also augments the input sign by pragmatic information, i.e. by information concerning its speech act.

The plan-based and the other two layers – statistics and finite state machine – interact in a number of ways: in cases where gaps occur in the dialogue statistical rating can help to determine the speech acts which are most likely to miss. Also, when the finite state machine detects an error, the planner must activate plan operators which are specialized for recovering the dialogue state in order not to fail. For this purpose specialized *repair-operators* have been implemented which determine both the type of error occurred and the most likely and plausible way to continue the dialogue. It is an intrinsic feature of the dialogue planner that it is able to process any input – even dialogues which do not the least coincide with our expectations of a valid dialogue – and that it proceeds properly if the parts processed by VERBMOBIL contain gaps.

## 4.2 The Statistical Layer – Statistical Modeling and Prediction

Another level of processing is an implementation of an information-theoretic model. In speech recognition *language models* are commonly used to reduce the search space when determining a word that can match a given part of the input. This approach is also used in the domain of discourse modeling to support the recognition process in speech-processing systems [Niedermair, 1992, Nagata and Morimoto, 1993]. The units to be processed are not words, but the *speech acts* of a text or a dialogue. The basis of processing is a training corpus annotated with the speech acts of the utterances. This corpus is used to gain statistical information about the dialogue structure, namely unigram, bigram and trigram frequencies of speech acts. They can be used for e.g. the prediction of following speech acts to support the speech processing components (e.g. dialogue dependent language models), for the disambiguation of different readings of a sentence, or for guiding the dialogue planner. Since the statistical model always delivers a result and since it can adapt itself to unknown structures, it is very robust. Also, if the statistic is updated during normal operation, it can adapt itself to the dialogue patterns of the VERBMOBIL user, leading to a higher prediction accuracy.

If we consider a dialogue to be a source that has speech acts as output, we can predict the $n$th speech act $s_n$ using the maximal conditional probability

$$s_n := \max_s \ P(s|s_{n-1}, s_{n-2}, s_{n-3}, ...)$$

We approximate $P$ with the standard smoothing technique known as deleted interpolation [Jellinek, 1990], using unigram, bigram and trigram relative frequencies, where $f$ are relative frequencies and $q_i$ are weights whose sum is 1:

$$P(s_n|s_{n-1}, s_{n-2}) = q_1 f(s_n) + q_2 f(s_n|s_{n-1}) + q_3 f(s_n|s_{n-1}, s_{n-2})$$

Given this formula and the required N-grams we can determine the $k$ best predictions for the next speech acts.

In order to evaluate the statistical model, we made various experiments. In the table below the results for two experiments are shown. Experiment TS1 uses 52 hand-annotated dialogues with 2340 speech acts as training corpus, and 41 dialogues with 2472 speech acts as test data. TS2 uses another 81 dialogues with 2995 speech acts as test data.

| *Pred.* | *TS1* | *TS2* |
|---|---|---|
| 1 | 40,65 % | 44,24 % |
| 2 | 60,19 % | 66,47 % |
| 3 | 73,92 % | 81,46 % |

Compared to the data from [Nagata and Morimoto, 1993] who report prediction accuracies of 61.7 %, 77.5 % and 85.1% for one, two or three predictions respectively, our predictions are less reliable. The main reason is, that the dialogues in our corpus frequently do not follow conventional dialogue behavior, i.e. the dialogue structure differs remarkably from dialogue to dialogue.

# 5   An Annotated Example

To get an impression of the functionality of the dialogue module, we will show the processing of three sentences which are part of an example dialogue which has a total length of 25 turns. This dialogue is part of a corpus of 200 dialogues which are all fully processed by our dialogue component. Prior to sentence DE004 given below EL initialized the dialogue requesting a date for a trip[2].

---

[2]DE indicates the German speaker, VM the translation provided by VERBMOBIL and EL the English speaker.  # indicates pressing or release of the button that activates VERBMOBIL.

```
      DE004:    #oh ja, gut, nach meinem Termin-
                kalender <Pause>, wie w"ars im
                Oktober?# (VORSCHLAG)
      VM005:    just lookin at my diary,  I would
                suggest October. (VORSCHLAG)
      DE006/1:  <Pause> I propose from  Tuesday
                the fifth/-
      DE006/2:  <Pause> no, Tuesday the fourth to
                Saturday the eighth <Pause>,
                those five days? (VORSCHLAG)
      EL007:    oh, that's too bad, I'm not free
                right then. (ABLEHNUNG) <Pause>
                I could fit it into my schedule
                <Smack> the week after, from
                Saturday to Thursday, the
                thirteenth. (VORSCHLAG)
```

If we trace the processing with the finite state machine and the statistics component, allowing two predictions, we get the following results:

```
        EL003:      INIT_TERMINABSPRACHE
        Prediction: (VORSCHLAG AUFFORDERUNG_VORSCHLAG)
        DE004:      VORSCHLAG
        Prediction: (AKZEPTANZ VORSCHLAG)
        DE006/1:    VORSCHLAG
        Prediction: (AKZEPTANZ VORSCHLAG)
        DE006/2:    VORSCHLAG
        Prediction: (AKZEPTANZ VORSCHLAG)  **Failed**
        EL007/1:    ABLEHNUNG
        Prediction: (VORSCHLAG AUFFORDERUNG_STELLUNG)
        EL007/2:    VORSCHLAG
        Prediction: (AKZEPTANZ ABLEHNUNG)
```

While the finite state machine accepts the sequence of speech acts without failure the predictions made by the statistical module are not correct for DE006/2. The four best predictions and their scores are AKZEPTANZ (28.09%), VORSCHLAG (26.93%), ABLEHNUNG (21.67%) and AUFFORDERUNG_STELLUNG (9.7%). In comparison with the fourth prediction, the first three predictions have a very similar ranking, so that the failure can only be considered a near miss. The overall prediction rates for the whole dialogue are 56.52 %, 82,60%, and 95.65% for one, two, and three predictions, respectively.
Since the dialogue can be processed properly by the finite state machine no repair is necessary. The only task of the planner therefore is the construction of the dialogue memory. It adds the incoming speech acts to the intentional structure, keeps track of the dates being negotiated, stores the various linguistic realizations of objects (e.g. lexical variations, referring expressions) and builds and administrates the links to the instantiated representation of these objects in the knowledge representation language BACK [Hoppe et al., 1993]. In fig. 3 we give two snapshots showing how the dialogue memory looks like after processing the turns DE006/2 and EL007.

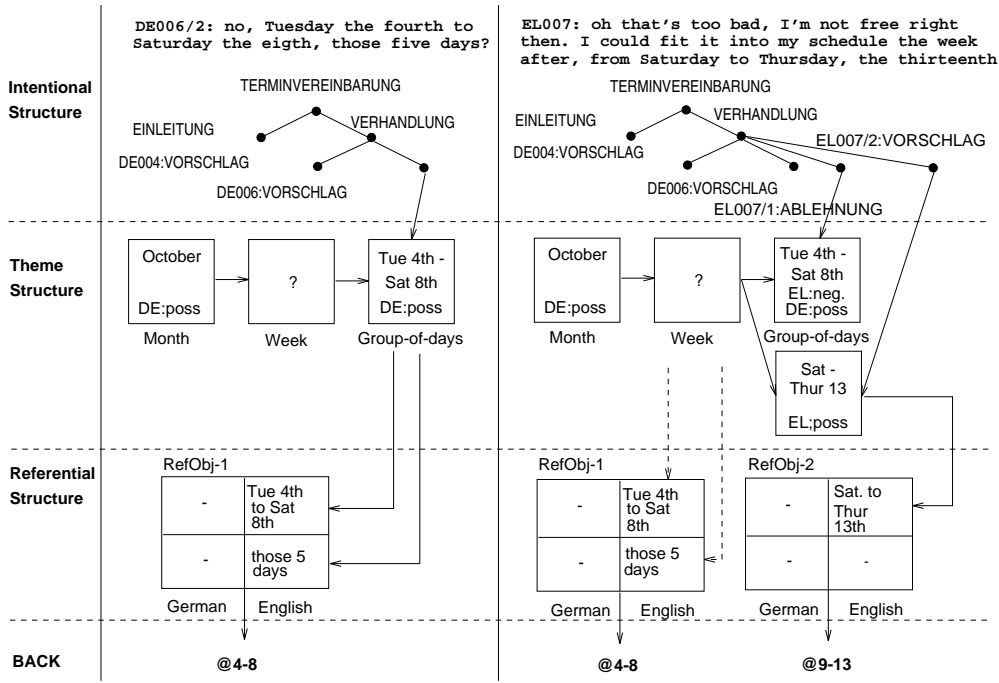

Figure 3: A snapshot of the dialogue memory after processing the utterances DE006/2 and EL007

# 6   Conclusion and future extensions

Dialogue processing in VERBMOBIL poses problems that differ from other systems like [Mast, 1993] and [Bilange, 1991]. Not being in a controlling position within a speech-processing system but tracking a mediated dialogue calls for an architecture where different approaches to dialogue processing cooperate. One important goal of our module is to provide top-down information for the other modules of VERBMOBIL, e.g. to reduce the search space of the word recognizer. This requirement is solved partially by using a statistics-based speech act prediction component. Also, we represent contextual information that is important for other VERBMOBIL components, as e.g. transfer and generation. This information is built up by the planner during dialogue processing.

Future extensions of the dialogue component, which has been sucessfully tested with 200 dialogues of our corpus concern the treatment of clarification dialogues. Robust processing will be another issue to be tackled: the possibility to process gaps in the dialogue will also be integrated.